\documentclass{article}

 \usepackage[nonatbib, preprint]{neurips_2020}

\usepackage{booktabs} 
\usepackage{graphicx, multirow} 
\usepackage{subcaption}
\usepackage[utf8]{inputenc} 
\usepackage[T1]{fontenc}    
\usepackage{url}            
\usepackage{booktabs}       
\usepackage{amsfonts}       
\usepackage{nicefrac}       
\usepackage{microtype}      
\usepackage{amsthm}
\RequirePackage[colorlinks,citecolor=blue,urlcolor=blue]{hyperref}
\usepackage{algorithm,algpseudocode}
\RequirePackage{amsmath, subcaption}
\RequirePackage{amsmath,dsfont}
\usepackage{amsthm}
\usepackage{amssymb,accents}
\usepackage{graphicx}
\usepackage{amssymb}
\usepackage{enumitem}
\usepackage{ulem}
\usepackage{graphicx}
\usepackage{geometry}
\usepackage{microtype}
\usepackage{graphicx}
\usepackage{booktabs} 
\usepackage[font=small,labelfont=bf]{caption}
\usepackage{bbm}

\newcommand{\dest}{\textit{destination} }
\newcommand{\source}{\textit{source} }
\newcommand{\posBias}{\textrm{posBias}}

\newcommand{\destUtility}{U_{\text{dest}}}
\newcommand{\sourceUtility}{U_{\text{source}}}

\newcommand{\aff}{\textrm{Aff}}
\newcommand{\graphAff}{\textrm{Aff}_{\textrm{Graph}}}

\input{definitions}

\title{A Framework for Fairness in Two-Sided Marketplaces}

\author{%
  Kinjal Basu, Cyrus DiCiccio, Heloise Logan, Noureddine El Karoui\\
  LinkedIn Corporation\\
  Mountain View, CA 94043\\
  \texttt{\{kbasu, cdiciccio, hlogan, nkaroui \} @linkedin.com} \\
}

\begin{document}

\maketitle

\begin{abstract}
Many interesting problems in the Internet industry can be framed as a two-sided marketplace problem. Examples include search applications and recommender systems showing people, jobs, movies, products, restaurants, etc. Incorporating fairness while building such systems is crucial and can have deep social and economic impact (applications include job recommendations, recruiters searching for candidates, etc.). In this paper, we propose a definition and develop an end-to-end framework for achieving fairness while building such machine learning systems at scale. We extend prior work \cite{joachims2018} to develop an optimization framework that can tackle fairness constraints from both the \source and \dest sides of the marketplace, as well as dynamic aspects of the problem. The framework is flexible enough to adapt to different definitions of fairness and can be implemented in very large-scale settings. We perform simulations to show the efficacy of our approach.
\end{abstract}

\vspace{-0.2cm}
\section{Introduction}
\label{sec:intro}
\vspace{-0.2cm}
Most early work on fairness in machine learning has focused on classification tasks and has strived to achieve fairness as defined through notions such as demographic parity, equality of opportunity, and equalized odds on protected attributes \cite{pmlr-v80-agarwal18a, Hardtetal_NIPS2016}.  Increasingly, businesses are relying on AI to recommend goods and services, thereby creating a need to define and address the bias that may exist in large-scale ranking or recommender systems. The traditional definitions of fairness do not necessarily extend to ranking or recommender systems, and sensible definitions of fairness in such systems are inherently more challenging than the classification use case \cite{joachims2018}. These challenges are magnified when people are both viewing recommendations from as well as being recommended by the ranking system \cite{pmlr-v81-burke18a,Mehrotra2018}, presenting a need for fairness to members engaging with a recommender through multiple roles. 


Depending on the application, recommender systems may be facilitating either a single-sided or multi-sided marketplace. A single-sided marketplace, in which users participate through a single role, usually consists of people who are either consumers or viewers, (for example members viewing items \cite{linden2003amazon}, products \cite{schafer1999recommender}, jobs \cite{Kenthapadi:2017:PJR:3109859.3109921}, movies, restaurants \cite{park2008restaurant}, etc.). A two-sided member-to-member marketplace, which arises when members can serve multiple functions, consists of the interplay between two parties such as consumers and producers, or viewers and viewees.  While the terminology used to describe the two parties is often application dependent, for simplicity, we will refer to the consumer or viewer side as the \textbf{\source} side and the producer or viewee side as the \textbf{\dest} side.  Common examples of two-sided marketplaces include feed ranking \cite{Agarwal:2014:ARL:2623330.2623362, Agarwal:2015:PLF:2783258.2788614}, people or friend recommendations \cite{pymk} and search systems \cite{agarwal_chen_2016} involving people on both sides such as recruiters searching for job candidates \cite{Borisyuk:2017:LSJ:3097983.3098028}. Moreover, recommender systems can be categorized according to whether they require explicit queries (for example people searching of items, restaurants, etc., recruiters searching for candidates, and so on) or implicit queries (users visiting to a platform to look at feed posts, check friend or connection recommendations, etc.).   


 


We are interested in guaranteeing fairness to groups of participants engaging in two-sided marketplace products. To ensure such a product is treating members fairly, especially with regards to protected attributes, we want to guarantee that the rankings are fair to the \source members initiating the queries (either explicit or implicit), as well as the \dest members who are being returned by the query.  

In this paper, we first describe formal notions of fairness using member utility in a two-sided marketplace. The notions of utility we use in this paper are related to fairness in rankings \cite{asudehDesigningFairRankingSchemes2019, BiegaEquityOfAttention2018, celis2017ranking, joachims2018, ZehlikeFAIRRanking2017}. Based on these definitions, we cast the problem of a fair two-sided marketplace as an optimization problem subject to appropriate fairness constraints. We extend the work of Singh and Joachims \cite{joachims2018} that considers (in our
terminology) the \dest-side fairness on a per-session, per-query level, in the following ways: 
\begin{itemize} [noitemsep,topsep=0pt,parsep=0pt,partopsep=0pt]
\item We define a multi-session utility and add fairness constraints for \dest members across multiple sessions
\item Through a similar setup, we add fairness constraints for \source members.  
\end{itemize}
The extension to multi-session fairness requires a significant reformulation of the fairness criteria; however, we still manage to keep the optimization formulation linear even after adding the different kinds of fairness constraints described above. Under this new framework, we provide a re-ranker, which is fair to both \source and \dest members, while transitioning from a single-session to a multiple-session framework. We also provide a solution that can be deployed online (as opposed to a static solution) and scale for large-scale recommender systems by using a duality idea.  

The rest of the paper is organized as follows. In Section \ref{sec:problem} we define the notions of fairness and formally state the problem. Section \ref{sec:destFairness} describes the fairness constraints, and the optimization framework for \dest members. We extent the procedure to \source members in Section \ref{sec:sourceSide}. Simulation experiments are shown in Section \ref{sec:experiment} to show the efficacy of our approach, before concluding with a discussion in Section \ref{sec:conclusion}. We end this section with an overview of the related works in literature.




\textbf{Related Work: }
In general, there are three approaches to debiasing in a recommendation system.  The first approach is preprocessing which aims to remove bias that may exist in the training data \cite{Kamiran2009ClassifyingWD,LuongKDD2011} prior to learning a model.  Since features may be correlated with protected attributes, it may be difficult to tease out the biasing signal completely.  
The second approach is to incorporate debiasing into the learning algorithm itself \cite{KamishimaPenalized2012,bechavod2017penalizing,pmlr-v54-zafar17a,pmlr-v28-zemel13}.  This approach is highly dependent on model architecture and typically cannot easily scale when the choice of modeling changes.  The third approach is through postprocessing on the final ranking produced by the recommendation system \cite{GeyikKDD2019,joachims2018}.  This approach is often favored as it is agnostic to the underlying model architecture. Our proposed methodology is a postprocessing approach which operates on the final ranking with fairness constraints, and is therefore very generally applicable.


Most of the work for a fair marketplace focuses on achieving fairness for the \dest side. \cite{GeyikKDD2019,joachims2018}. In a talent search product through which a recruiter searches for qualifying candidates, the recommendation system strives to achieve fairness on the candidate side by guaranteeing a desired distribution of top ranked results with respect to gender \cite{GeyikKDD2019}.  Singh and Joachim's work \cite{joachims2018} introduces certain notions of fairness for \dest members and focuses on maximizing utility with fairness constraints via a linear programming setup. Burke's work \cite{pmlr-v81-burke18a} talked about multi-sided fairness which is described as consumer side and producer side fairness.  The balanced neighborhoods approach ensures fairness from one side at a time.  Mehrotra's work \cite{Mehrotra2018} on music recommendation service is one of the few that considers both source and destination side fairness.  
They proposed several policies to tradeoff relevance and fairness, and have a counterfactual estimation framework to evaluate these policies offline. 

However, a fully generic framework for fairness in a two-sided marketplace is missing in the literature, and in this paper, we try to fill that gap by extending the work from \cite{joachims2018}. Fairness in a two-sided marketplace can be conceptually considered as a problem on an undirected graph, whereas previous work has implicitly been considering bipartite graph situations \cite{GeyikKDD2019,joachims2018}. This point of view does not seem to have received as much attention in the literature as the more classical question of fairness in classification, though see \cite{pmlr-v81-burke18a,Mehrotra2018}.




\vspace{-0.2cm}
\section{Notions of Fairness in a Two-Sided Marketplace}
\label{sec:problem}
\vspace{-0.2cm}

In much of the fairness literature (e.g \cite{Hardtetal_NIPS2016,mary2019fairness}), fairness can be understood as a conditional independence requirement.  Suppose that $U(m)$ represents the utility that a user $m$ has for a system (which is usually a univariate model score).  Suppose that members belong to one of several groups, $G_1,...,G_K$ (for instance, group membership may be determined by a protected attribute such as gender or race).  Fairness can be defined as an independence condition between the utilities received by members and group membership, potentially subject to conditions on member attributes.  Formally, this requirement can be formulated as, if we write $m \in G_k$ for membership to group $G_k$,
\begin{align*}
\textrm{Law}\, (U(m) | m \in G_k,Z) = \textrm{Law} (U(m) | m \in G_{k'},Z)\;, \forall k,k'\;
\end{align*}
 where $Z$ is a random variable encompassing an additional layer of conditioning. Concretely, with equalized odds (resp. equality of opportunity) \cite{Hardtetal_NIPS2016}, $Z$ is a label and one conditions on $Z=0$ or $Z=1$ (resp $Z=1$). 
 Many notions of fairness are derived from these equality in law requirements by using a summary statistic of the distribution such as the mean  (e.g. \cite{optimizedScoreTransfoduPinCalmon19}).

However, in a two-sided marketplace every participant in the market has fundamentally a 2-dimensional vector of utility: one aspect of it is the utility they get as an active participant (we call this the \textbf{source} utility, e.g. the utility derived from issuing a query) and the other is the utility they get as a passive participant (we call this the \textbf{destination} or simply \textbf{dest} utility, e.g. the utility derived from being recommended in a recommender system). Fairness notions, seen as conditional independence requirements, are therefore easily extended to this 2-sided marketplace setting by  requiring equality of the joint distribution of these bivariate utilities. Practically, this condition can be relaxed by simply requiring equality of the marginal distributions of these utilities. 
\begin{definition}\label{def:2SidedFairness} Fairness from the \source and \dest side in a 2-sided marketplace can be written as: for all $k,k'$,
\begin{align*}
\textrm{Law}\, (\sourceUtility(m) | m \in G_k,Z) &= \textrm{Law} (\sourceUtility(m) | m \in G_{k'},Z)\;\; \text{and}\\
\textrm{Law}\, (\destUtility(m) | m \in G_k,Z) &= \textrm{Law} (\destUtility(m) | m \in G_{k'},Z)\;. 
\end{align*}
\end{definition}
In practice it might be convenient to get a single summary statistic for the total marketplace utility, which gives rise to other simplified notions of fairness. At the distributional level, one could consider equality of the conditional distributions of $\sourceUtility(m)+\lambda \destUtility(m)$ for a pre-specified $\lambda$ for instance. Another layer of simplification comes from moving from distributional requirements to further ``projections'' or summary statistics, such as moments; the mean is then very natural. 

Throughout the remainder of the paper, we will consider fairness conditions which are moment-based approximations to the conditional distribution requirements, working off of Definition \ref{def:2SidedFairness}.  These choices ensure that a suitable reranker both exists and can be implemented in large scale settings.  
\vspace{-0.1cm}
\paragraph{Formal problem statement: } We will begin with some notational conventions before introducing the problem formally.
Let $s$ denote the \textit{source} member initiating a query $q$ and $d$ denote a \textit{destination} member being ranked as a result of the query.  Let $u_{s,d}^q$ denote the final ranking score for this pair, which reflects the utility that member $s$ receives from being shown member $d$.  Let $P_s^q(d,r)$ denote the probability of showing the $d$-th destination member in the $r$-th slot for source $s$ and query $q$.  Finally, let $v_r$ denote the exposure of the $r$-th slot. We are interested in providing a ranking of the top $m$ destination members, where $m < D_q$, the total number of eligible \dest members for query $q$. Thus, the ranking matrix $P_s^q$ is of dimension $D_q \times m$.

In this setting, the expected source side utility, for each $(s,q)$ can be defined as
\begin{align}
\vspace{-0.1cm}
\label{eq:sourceUtility}
U_q^{\textrm{source}}(s) &= \sum\nolimits_{d = 1}^{D_q} \sum\nolimits_{r =1}^m u_{s,d}^q  P_s^q(d,r)  v_r  =  u_s^\top P_s v
\end{align}
where $u_s$, $P_s$, and $v$ are the appropriate vectors and matrices. 
\begin{remark}
Note that the position bias $v$ does not usually depend on the query $q$. Moreover, for notational simplicity we remove the superscript $q$ unless explicitly required.
\end{remark}

Most recommender systems and marketplace problems \cite{agarwal_chen_2016} try to maximize this source side utility. Destination side utility is much harder to tackle, since it depends of different queries arising from different sources. The expected utility a \dest member $d$ receives from being shown to member $s$ at rank $r$ for a query $q$ can be written as, 
\begin{align}
\label{eq:destUtility}
U_q^{\textrm{dest}}(s,d) = u_{s,d}^q \sum\nolimits_r P_s^q(d,r) \cdot v_r = u_{s,d}^q P_s^q(d, \cdot) v 
\end{align}
Define for the destination member $d$, who has been shown to members $s_{t_1},...,s_{t_N}$ from queries $q_{t_1},...,q_{t_N}$ at times ${t_1},...,{t_N}$ prior to time $T$, the utility up to time $T$ as 
\begin{align}
\vspace{-0.2cm}
\label{eq:destFinalUtility}
U^{\textrm{dest}}(d)[T] = \sum\nolimits_{i=1}^{N} \rho^{T - t_i}  U_{q_{t_i}}^{\textrm{dest}}(s_{t_i},d)
\end{align}
where $\rho$ is an appropriately chosen discount rate.  Explicitly, this definition assumes that older sessions are less valuable than more recent sessions,  unless $\rho=1$ in which case all are equally valuable.  We will show in Section \ref{sec:dynamic} how this definition is used in our framework.

\begin{remark}
One important difference between the source and destination side utilities is the fact that for source-side utility for a member can be estimated at a single session-level, while destination side utility is across sessions. This difference stems from the objective of the optimization problem being defined to maximize the session-level source side utility. We will, however, see in Section \ref{sec:sourceSide} that source-side utility can also be handled across sessions and address how to account for multiple sessions. 
\end{remark}

By introducing an optimization-based framework \cite{joachims2018} tackle the problem of maximizing source side utility while maintaining fairness constraints on the \dest members. Naturally, the constraints are at a per-session level. We expand on their framework by tackling two challenging problems: 
\begin{description}
\item[Problem 1:] Adding fairness constraints for \dest members at multiple sessions by considering the multi-session utility as given in \eqref{eq:destFinalUtility}.
\item[Problem 2:] Adding fairness constraints for \source members.  
\end{description}

We first focus on solving Problem 1. Extending the optimization based framework from \cite{joachims2018}, the overall problem of fairly ranking \dest members can be written as maximizing the source side utility for member $s$ and query $q$ while keeping the allocation of the destination members fair:
\begin{equation}
\begin{aligned}
    \label{eq:opt}
    \mbox{Maximize }& u_s^\top P_s v \\
    \mbox{Subject to } &1^\top P_s = 1,  \;\;\; P_s 1 \leq 1, \;\;\; 0 \leq P_s(d,r) \leq 1, \;\;\;\text{and}\;\;\; P_s \mbox{ is fair}.
\end{aligned}
\end{equation}

The first constraint enforces the condition that in each slot $r$ we must show a \dest member, while the second constraint ensures that each \dest member may or may not be shown ($m < D_q$). In the case when $m = D_q$, we get equality in the second constraint ($P_s$ becomes constrained to be a doubly stochastic matrix) and the problem reduces to the exact framework of \cite{joachims2018}. 

Several notions of fairness have been discussed in \cite{joachims2018}, all of which are applicable here. Moreover, we show that fairness constraints with respect to the multi-session \dest utility \eqref{eq:destFinalUtility} can also be incorporated in a very similar fashion, while still solving the single source member and single query optimization problem as given in \eqref{eq:opt}. In general, if the fairness constraint or other constraints are linear, the problem remains a linear program in $P_s$ which can be solved either offline or online. We will discuss the scalability aspects of this problem in Section \ref{sec:opt}. We primarily focus on solving Problem 1 in a scalable manner under fairness constraints. We briefly describe how we can also solve Problem 2 as an extension. The reason being that Problem 2 can be often at direct odds to what the marketplace or the recommender system is aiming to achieve. We discuss this in detail in Section \ref{sec:sourceSide}. 

\vspace{-0.2cm}
\section{Destination Side Fairness}
\label{sec:destFairness}
\vspace{-0.2cm}
Singh and Joachims \cite{joachims2018} cover a wide range of notions of \dest member fairness, all of which are applicable in our setting. Following a similar idea, we also consider fairness with respect to disjoint member groups $G_1, \ldots, G_k$. For completeness, we briefly describe some of the notions here.

\textbf{Demographic Parity: } A cornerstone of fairness to the \dest members being shown for a query is making certain that various groups receive fair exposure in the rankings. Mathematically, this can be expressed as \[
\frac{1}{|G_k|} \sum\nolimits_{d \in G_k} \sum\nolimits_r P_s^q(d,r) v_r = \frac{1}{|G_{k'}|} \sum\nolimits_{d \in G_{k'}} \sum\nolimits_r P_s^q(d,r) v_r
\]
for all $k,k'$.  More compactly, these constraints can be written as $f^\top_{\textrm{DP}, k,k'} P_s v = 0$, where the $d$-th component of $f_{\textrm{DP}, k,k'}$ is 
$\frac{\mathbbm{1}_{d \in G_k}}{|G_k|} - \frac{\mathbbm{1}_{d \in G_{k'}}}{|G_{k'}|}$.

\textbf{Disparate Treatment:} This criteria requires that the ratio of exposure of the destination members being ranked to the utility that they provide be equal across groups: 
\[
\frac{1}{|G_k| \bar U_q(G_k)} \sum\nolimits_{d \in G_k} \sum\nolimits_r P_s^q(d,r) v_r = \frac{1}{|G_{k'} |\bar U_q(G_{k'})} \sum\nolimits_{d \in G_{k'}} \sum\nolimits_r P_s^q(d,r) v_r
\]
where $\bar U_q(G_k) = \sum_{d \in G_k } u_{s,d}^q / |G_k|$. Similar to above, the constraint reduces to $f^\top_{\textrm{DT}, k,k'} P_s v = 0$, where the $d$-th component of $f_{\textrm{DT}, k,k'}$ is 
$\frac{\mathbbm{1}_{d \in G_k}}{|G_k|\bar U_q(G_k)} - \frac{\mathbbm{1}_{d \in G_{k'}}}{|G_{k'}|\bar U_q(G_{k'})}.
$

\textbf{Disparate Impact:} This criteria is similar to disparate treatment but explicitly uses the impact of the ranking based on the utility instead of the exposure on the numerator. Formally,
\[
\frac{1}{|G_k| \bar U_q(G_k)} \sum\nolimits_{d \in G_k} u_{s,d}^q \sum\nolimits_r  P_s^q(d,r) v_r = \frac{1}{|G_{k'} |\bar U_q(G_{k'})} \sum\nolimits_{d \in G_{k'}} u_{s,d}^q \sum\nolimits_r P_s^q(d,r) v_r
\]
which reduces to $f^\top_{\textrm{DI}, k,k'} P_s v = 0$, where the $d$-th component of $f_{\textrm{DI}, k, k'}$ is 
$
\frac{u_{s,d}^q\mathbbm{1}_{d \in G_k}}{|G_k|\bar U_q(G_k)} - \frac{u_{s,d}^q\mathbbm{1}_{d \in G_{k'}}}{|G_{k'}|\bar U_q(G_{k'})}$.

\vspace{-0.2cm}
\subsection{Dynamic and Multi-Session Fairness}
\label{sec:dynamic}
\vspace{-0.2cm}
It is important to ensure that members are receiving utility fairly across multiple sessions.  In some instances, we can measure a member's utility over time, and ensure that groups are being treated fairly - dynamically (see also \cite{damour20}). However, a significant conceptual challenge arises: dynamic fairness is fundamentally a multi-session problem, whereas our optimization problem is framed on a per-session basis. Using the \dest member utility $U^{\textrm{dest}}(d)[T]$ as defined in \eqref{eq:destFinalUtility} we describe how a multi-session fairness constraint can be defined and incorporated in the per-session optimization problem, for example, by requiring that the average cumulative utility be equal between two groups.

\begin{definition} The average cumulative utility for group $G_k$ at time $t$ is 
$$\mu_{G_k}[t] = \frac{1}{|G_k|}\sum\nolimits_{d \in G_k} U^{\textrm{dest}}(d)[t].$$
\end{definition}
If the next query occurs at time $T$, it easily follows from the \eqref{eq:destFinalUtility} that,
\begin{align}
\label{eq:destupdate}
U^{\textrm{dest}}(d)[T] = U_{q_T}^{\textrm{dest}}(s_T,d) + \rho^{T-t} U^{\textrm{dest}}(d)[t].
\end{align}
Note that, maintaining the destination side utility with respect to time is computationally efficient due to the simple update equation as given in \eqref{eq:destupdate}. Moreover, the incremental utility can be written as 
\begin{align*}
\Delta \mu_{G_k}[T] &= \mu_{G_k}[T] - \mu_{G_k}[t] = \frac{1}{|G_k|} \sum\nolimits_{d \in G_k}U_{q_T}^{\textrm{dest}}(s_T,d) + (\rho^{T-t} - 1) \mu_{G_k}[t]\\
&= \frac{1}{|G_k|} \sum\nolimits_{d \in G_k} \sum\nolimits_r u_{s_T,d}^{q_T} P_{s_T}^{q_T}(d,r) v_r + (\rho^{T-t} - 1) \mu_{G_k}[t]
\end{align*}

Consequently, a fairness constraint requiring equality of average incremental utility across groups,  $\Delta \mu_{G_k}[T] = \Delta \mu_{G_{k'}}[T]$
for each $k$ and $k'$ can be written as 
$\tilde u_s P_s w = c$
where $\tilde u_s$ is a vector having entries 
$
u_{s,d}^q\left(\frac{\mathbbm{1}_{d \in G_k}}{|G_k|} - \frac{\mathbbm{1}_{d \in G_{k'}}}{|G_{k'}|}\right)
$
and $c = (1 - \rho^{T-t}) (\mu_{G_k}[t] - \mu_{G_{k'}}[t])$. Maintaining the cumulative utility for each member over time allows the rankings obtained by solving a linear program per each individual session and query to account for fairness with respect to multi-session metrics. 


\vspace{-0.2cm}
\subsection{Optimization Framework}
\label{sec:opt}
\vspace{-0.2cm}
Based on the above discussion, for any member $s$, the overall (source-utility) optimization problem while considering fairness metrics from the destination side can we written as 
\begin{equation}
\begin{aligned}
\label{eq:fullopt}
    \mbox{Maximize }& u_s^\top P_s v  \\
    \mbox{Subject to } &1^\top P_s = 1,  \;\;\; P_s 1 \leq 1, \;\;\; 0 \leq P_s(d,r) \leq 1, \;\;\; f_{k,k'}^\top P_s w = 0 \;\;\; \text{ and } \;\;\;  \tilde u_s P_s w = c 
\end{aligned}
\end{equation}

This is a linear program in $P_s$, considering fairness metrics between groups $G_k$ and $G_{k'}$. We consider only two groups for notational simplicity, but additional constraints can be trivially added to solve the problem for multiple groups. Moreover, we only consider a generic $f_{k,k'}$, but can easily add multiple such constraints as discussed in Section \ref{sec:destFairness}.

The optimization problem \eqref{eq:fullopt} can be solved at a per-session level, whenever we want to serve recommendations or results to the member $s$ for query $q$. This is preferred when the set of queries can be very diverse, for example, in search systems such as recruiters searching for candidates, people searching for member cohorts, etc.  However, for many other recommender systems, such as news feed ranking, friend recommendations, movie recommendations, etc... there is no inherent notion of a query  In such use cases, due to a large number of query-per-second (QPS), solving an optimization problem at an individual session level can be extremely challenging in industrial practice due to latency and storage considerations. Instead, we propose a mechanism to solve these at an aggregated session level (while still solving it for each member), thereby guaranteeing that the fairness constraint will hold across sessions as long as the score distribution of individual models does not differ drastically.

Let us assume we consider the last $S$ sessions for each member $s$. Let $M_s$ be the union of the set of all destination members that were shown in those sessions. Then, $P_s$ is of dimension $M_s \times m$ where $m$ is the total number of slots we have. Let $u_{s,d}$ denote the average utility of $(s,d)$ pair across the sessions. Then, the optimization problem is the same as \eqref{eq:fullopt} but across multiple sessions with the fairness constraints taken across the set of $M_s$ members seen throughout the $S$ sessions (rather than only the candidates for the current session). Note that, under this regime of multiple sessions or queries for source member $s$, the entities $u_s$ and $P_s$ are no longer dependent on the query. Analogous multi-session \dest member utilities can be similarly defined.

\paragraph{Solving the Optimization Problem: }
Since we solve the problem for each member $s$, we drop the notation $s$ for simplicity. Rewriting the optimization problem in a vectorized form, we have
\begin{equation}
\begin{aligned}
\label{eq:fulloptvector}
    &\mbox{Max } p^\top(u \cdot v)   \\
    &\mbox{s.t.}  \;\;\; p^\top (f_{k,k'} \cdot v) = 0, \;\;  p^\top (\tilde{u} \cdot v) = c, \;\; [I_m : I_m : \cdots : I_m] p = 1 \;\; \text{and}\;\; p_d \in T_m \;\; \forall \; d \in M 
\end{aligned}
\end{equation}
Here, $p = (p_1, \ldots, p_d, \ldots, p_M)$ is the vectorized version of the matrix $P$ of dimension $M \times m$, $( \cdot )$ denotes element-wise vector multiplication. $I_m$ is the identity matrix of size $m \times m$ and $T_m$ is the simplex of dimension $m$, i.e. $T_m = \{ (a_1,\ldots, a_m) | \sum_{r =1}^m a_r \leq 1, a_r \geq 0 \}$. We add a small regularizer to the objective function and solve the following optimization problem,
\begin{align}
\label{eq:fulloptvectorfinal}
    \mbox{Maximize }& p^\top(u \cdot v) + \frac{\gamma}{2} p^\top p  
    & \mbox{subject to } \;\;\; Ap \leq b \;\;\; \text{ and } \;\;\; p_d \in T_m \;\;\; \forall \; d \in M 
\end{align}
where $A,b$ encompasses the constraints as given in \eqref{eq:fulloptvector}. After solving the optimization problem we obtain the dual variables $\lambda_1, \lambda_2$ (corresponding to the fairness constraints) and $\eta_{m\times1}$ (corresponding to the slots/positions). We can then use these dual variables to obtain the primal solution in a online system, which drastically reduces the latency cost in serving such a system.

\paragraph{Dual To Primal Trick: }
Using the KKT conditions, it is fairly straightforward to obtain the unique primal solution of \eqref{eq:fulloptvectorfinal} as a function of the dual \cite{basu_eclipse, boyd2004convex}. Specifically, we have
\begin{align}
\label{eq:dualproj}
\hat{p}_d (\lambda_1, \lambda_2, \eta) = \Pi_{T_m} \Big( \frac{1}{\gamma} \left\{(u \cdot v)_d - \lambda_1(f_{k,k'} \cdot v)_d  - \lambda_2(\tilde{u} \cdot v)_d    - \eta \right\} \Big) 
\end{align}
where $\Pi_{T_m}$ denotes the projection onto the simplex $T_m$ and $(\cdot)_d$ denotes the vector component corresponding to the $d$-th member. Note that the primal solution only depends on the $m+2$ dual variables ($\lambda_1, \lambda_2$, and $\eta_{m \times 1}$).
Hence, when we have to display the results of the search or the recommended rankings for the member $d$, we evaluate the components of the corresponding vectors and perform the projection operation as shown in \eqref{eq:dualproj}. There are efficient algorithms for the simplex projection \cite{efficientProj} 
 which makes online scoring highly efficient. 


We apply the above transformation on each of the candidate members $d$ to obtain $\hat{p}_d$, for $d = 1, \ldots, D$ where $D$ denotes the total set of candidates to score. In almost all applications $D > m$ (i.e. it is not necessary to rank all $D$ candidates) and consequently, the obtained matrix $\hat{P}_{D\times m} = ( \hat{p}_1 : \hat{p}_2: \ldots: \hat{p}_D)^T$ is not a doubly stochastic. 
As a result, to get the final deterministic serving scheme we cannot use the approach of from \cite{joachims2018} via the Birkhoff-von Neumann decomposition \cite{birkhoff1940lattice}. Instead we use a simple greedy approach to generate the member at slot $r$ which also helps in reducing online latency:
\begin{align*}
\textrm{member}(r) = \argmax_{d : d  \textrm{ is not yet assiged}} \hat{P}(d,r) \;\;\; \text{for } r = 1, \ldots, m.
\end{align*}
Moreover, based on the chosen serving scheme we continuously update the destination side utility as defined in \eqref{eq:destupdate}. Before each run, we can tweak the \dest side fairness constraint to get any distribution on the destination side utility. See appendix for further discussion.



\vspace{-0.2cm}
\section{Extension to Source Side Fairness}
 \label{sec:sourceSide}
 \vspace{-0.2cm}
While the optimization problem \eqref{eq:opt} yields rankings which maximize the utility for the \source side members subject to providing fairness to the \dest members, it is naturally also of interest to ensure that the utility obtained by the source is not biased for one group versus the other. 

It is important to note the distinction here between \source and \dest side fairness. The \dest side unfairness is easily observable from a single \source member's point of view.  For instance, if a query results in large exposure of one group versus another, it can be easily seen and identified by the viewer (or \source member). On the contrary, if a group of source members is given consistently better results compared with another, it is not directly observable from the single source perspective. Nevertheless, since the ranking algorithms are by default trying to optimize the source side utility, it is unlikely that a group of source members would be consistently achieving poorer utility in a well-functioning system. 
Moreover, if such groups exist, then a lot of effort would actually be put in place in industrial practice to improve their utility through iterations of better machine learning models, as such efforts have a direct impact on business metrics. 
For this reason, our primary focus has been on \dest side fairness. Nevertheless, our framework is rich enough to handle multi-session source-side utility fairness constraints. We now describe how to do just that.

A baseline measure for \source side fairness is ensuring that they receive comparable expected utility across groups over multiple sessions. Suppose that members $s_1,...,s_{n_k}$ belonging to group $G_k$ initiate queries $q_1,...,q_{n_k}$ at times $t_1,...,t_{n_k}$, then we can define cumulative expected utility for group $G_k$ up to time $T$ as
\[
\EE \left[ U^{\textrm{source}}(G_k,T) \right] = \sum\nolimits_{i: t_i \leq T} \rho^{T-t_i} U_{q_i}^{\textrm{source}}(s_i) =  \sum\nolimits_{i: t_i \leq T} \rho^{T-t_i} u_{s_i}^\top P_{s_i} v
\]
where $\rho$ is a discount factor for the time of the query. Similar to the multi-session \dest utility, it is readily seen that if the first query $q$ after time $T$ occurs at time $T+\delta$, and is initiated by member $s$, then
$
\EE \left[ U^{\textrm{source}}(G_k,T + \delta) \right] = u_{s}^\top P_{s} v  + \rho^\delta \EE \left[ U^{\textrm{source}}(G_k,T) \right].
$
Consequently, if member $s$ belonging to group $G_k$ initiates a query $q$ at time $T+\delta$ , a constraint requiring that the cumulative expected utility be equal between groups, i.e. $\EE \left[ U^{\textrm{source}}(G_k,T + \delta) \right] = \EE \left[ U^{\textrm{source}}(G_{k'},T + \delta) \right]$
for all $k,k'$, can be written as $u_{s}^\top P_{s} v = \EE \left[ U^{\textrm{source}}(G_{k'},T + \delta) \right] - \rho^{\delta} \EE \left[ U^{\textrm{source}}(G_k,T) \right]$, 
which is a set of linear constraints and hence easily incorporated in our LP framework. 

Note that, however, this constraint somewhat conflicts with the maximization objective of \eqref{eq:fullopt} and brings to light the important concept of losing accuracy in lieu of bringing in fairness \cite{pmlr-v54-zafar17a}. 
Here, we present the framework which makes incorporating such constraint feasible and scalable. To incorporate both \source side and \dest side fairness, the full optimization problem can be written as:
\begin{equation}
\begin{aligned}
\label{eq:fullBothopt}
    \mbox{Minimize }& \sum\nolimits_{k' \neq k} \left|u_s^\top P_s v - c_{k,k'}\right| &   \\
    \mbox{Subject to } &1^\top P_s = 1,  \;\;\; P_s 1 \leq 1, \;\;\; 0 \leq P_s(d,r) \leq 1, \;\;\; f_{k,k'}^\top P_s w = 0, \;\;\; \tilde u_s P_s w = c  
\end{aligned}
\end{equation}
where $c_{k,k'} = \EE \left[ U^{\textrm{source}}(G_{k'},T + \delta) \right] - \rho^{\delta} \EE \left[ U^{\textrm{source}}(G_k,T) \right]$ or an appropriately chosen value to sway the distribution in some direction. This is still a linear program and can be handled in a very similar manner as we discussed in Section \ref{sec:opt}.

\vspace{-0.2cm}
\section{Experimental Results}
\label{sec:experiment}
\vspace{-0.2cm}
To investigate the behavior of our proposals in a controlled environment, we simulate a member-to-member recommender system, which provides members with suggestions of other members with whom to ``connect'', i.e. a simple version of a two-sided marketplace. 
\vspace{-0.2cm}
\paragraph{Simulation setup:}
We use a graph to represent connections between members. Because we are interested in a two-sided marketplace where both ``sides'' of the marketplace play a similar role, our graph is not bipartite (as it would be in a consumer-product setting) but is a standard undirected graph. The vertices of the graph correspond to members and the edges to whether or not there is a connection between vertices. When we initialize the vertices, each vertex of the graph (representing a member) is given a binary group membership 
(on which we want to enforce fairness) 
and 
a vector of covariates, $X_i$. The member and graph affinity between two members $i$ and $j$ is measured as $\aff_{\textrm{member}}(i,j)=-\norm{X_i-X_j}_2$ and $\graphAff(i,j)=A_t^2(i,j)$, respectively, where $A_t$ is the adjacency matrix at time $t$, representing the number of common connections between $i$ and $j$.

The graph evolves in the following manner.  A vertex $i$ is picked at random. For each other vertex $j$, a ``model'' normalized score $s(i,j)$ is computed as a function of $\aff_{\textrm{member}}$ and $\graphAff$	. A subset of vertices of size $D$ with the highest scores are then considered eligible for showing. Among them, a smaller subset of size $m$ (possibly identified after re-ranking to incorporate fairness constraints, see below) is then ``shown''. For the subset of size $m$ of members that are shown, an edge (or connection) in the graph is created at random with a probability that decays according to the position. Please see the Appendix for further details. 
The fairness adjustments are dealt with in the following fashion: 
\begin{enumerate}[leftmargin=*] 
\setlength\itemsep{-0.2em}
\item No fairness adjustment in the ranking (\texttt{noReranker} in Table \ref{tab:results}) 
\item Equality of opportunity, using the primal for destination side fairness (\texttt{primal}) 
\item Equality of opportunity, using the dual method for destination side fairness (\texttt{dualNoDynamic})
\item Multi-Session destination utility adjustment via the dual (\texttt{dualWithDynamic}). 
\end{enumerate}

\vspace{-0.2cm}
\paragraph{Numerical results:} We keep track of the source and destination side utilities, which are aggregated by summation across sessions.  
We consider three simulation settings, each with 1000 members and 1000 iterations, but with different numbers of ``shown'' members, $m$, and different rates of the dual refresh, $t$ (the gap in iterations where we re-solve the primal). We report the average difference in exposure between the two groups ($\Delta$DP) and the mean absolute value difference for this quantity, the ratio of average source utility between the groups and the percentage of total destination utility in Group 0 as our metrics (see the appendix for precise definitions). The results for a single simulation is shown in Table \ref{tab:results}. For more detailed result with multiple runs and variance estimates, please see the appendix.

\begin{table}[!h]
\centering
\begin{small}
\begin{tabular}{|c|c|c|c|c|c|} 
 \hline
 \multirow{2}{*}{Settings} &\multirow{2}{*}{Method}& \multirow{2}{*}{$\Delta$DP} & \multirow{2}{*}{$\Delta |$DP$|$} & Ratio of & \% Total \\
 &  &  &  & Source Utility & Destination Utility \\ [0.5ex] 
 \hline\hline
&\texttt{noReranker} &-0.0216&0.1964&0.5100&0.7051 \\ 
$m = 10$&\texttt{primal}&0.0179&0.1671&0.5024&0.5371\\ 
$t = 50$&\texttt{dualNoDynamic} &0.0055&0.2125&0.5100&0.5860\\ 
& \texttt{dualWithDynamic}&0.0249&0.2008&0.5100&0.5706\\ 
 \hline\hline
&\texttt{noReranker}&-0.0072&0.1182&0.5100&0.7059\\ 
$m = 20$&\texttt{primal}&0.0109&0.1002&0.5023&0.5269\\ 
$t= 50$&\texttt{dualNoDynamic}&0.0122&0.1090&0.5099&0.5536\\ 
& \texttt{dualWithDynamic}&0.0155&0.1099&0.5099&0.5371\\ 
\hline\hline
& \texttt{noReranker} &-0.0072&0.1182&0.5100&0.7059\\ 
$m = 20$&\texttt{primal}&0.0109&0.1002&0.5023&0.5269\\ 
$t = 20$ &\texttt{dualNoDynamic}&0.0177&0.1105&0.5099&0.5210\\ 
&\texttt{dualWithDynamic}&0.0185&0.1090&0.5099&0.5166\\
\hline
\end{tabular}
\end{small}
\caption{Simulation Results}
\label{tab:results}
\vspace{-4mm}
\end{table}

\textbf{Learnings:} The simulations illustrate the fact that adding fairness re-rankers help rebalance \dest utility (last column) while not significantly altering \source utility metrics (see the 5th column in Table \ref{tab:results}). Refreshing the dual parameters more often helps bring \dest utility results closer (see last two rows and the last column in Table \ref{tab:results}), at the cost of slower computations (not reported here). Increasing the number of ``shown'' slots makes satisfying fairness constraints easier a priori, and this is also verified in our simulations (first two rows and last column). We did not implement the source utility re-ranker in this set of simulations since even without mitigation the setup was such that source utility was comparable across both groups (columns 5 above)- see also discussion in Section~\ref{sec:sourceSide}. We leave this to future work and real-world implementation.

\vspace{-0.2cm}
\section{Conclusion}
\label{sec:conclusion}
\vspace{-0.2cm}
Multi-sided marketplaces can involve intricate member to member interactions, presenting a challenge for defining fairness and ensuring fairness not only by groups of members, but also by the role in which they are participating in the marketplace. Existing literature has generally addressed the issue of fairness in a single-sided marketplace. Our present work provides definitions of fairness for two-sided marketplaces, 
and guarantees that members are given fair treatment over-time with respect to source and destination side utilities that are computed across sessions. 
Our simulations served as proof of concept to demonstrate that without proper intervention, two-sided marketplaces can present severe implicit discrimination - in terms of utility or representation/exposure -  against groups of underserved members. 
Moreover, the present framework can help limit over time such discrimination. 

\vfill
\pagebreak

\section*{Broader Impact}
Machine Learning (ML) systems may contain implicit biases and reproduce or reinforce them. It is hence crucial to develop  tools for assessment and mitigations of potential implicit biases in large scale ML systems, since they increasingly power decisions affecting larger and larger portions of society. 

The situation where the ML system powers a 2-sided marketplace has received less attention in the literature than the now classic question of fairness in classification. We tackle this fairness-in-a-2-sided marketplace problem by proposing tools grounded in sound theoretical principles and practically applicable at Internet-data scale. We hope that these tools will be broadly applicable where they matter crucially, in real-life applications. Our methods try to measure and mitigate unfairness with respect to classic ethical principles, rooted in moral and political Philosophy principles using for instance the concept of disparate impact (see also Equal Employment Opportunity Commission guidelines and rules) and disparate treatment. 

This research should benefit groups that are currently disadvantaged in large-scale 2-sided marketplaces powered by automatic recommendation systems and should yield positive outcomes for those groups. The consequences of failure of the system might be the perpetuation of the status quo in terms of performance (broadly construed) of those systems. We have tried to address questions related to the  dynamic aspects of fairness which remain very challenging. Our paper has not addressed intersectionality issues explicitly, though the framework is rich enough to handle some intersectional questions. Hence good practice will require not only monitoring the fairness performance of the tools we have developed but also whether they have unintended intersectional consequences and potentially refining some of the constraints in our LP formulation to mitigate them if they appear in practice.

\def\bibfont{\small}
\bibliography{fairness,biblioRelatedWork}
\bibliographystyle{abbrv}

\vfill
\pagebreak
\appendix
\section{Appendix}
\subsection{Further details on simulation experiments}\label{subsec:DetailsSimulations}
Here are more details on the simulations were discussed in Section \ref{sec:experiment}. 

\paragraph{Graph Initialization:} When we initialize the graph, we create a stochastic block model \cite{NewmanStochBlockModelPhysRevE.83.016107} with 2 blocks to initialize the edges. We call $A_0$ the corresponding adjacency matrix, with, by convention, $A_0(i,i)=0$ for all $i$. At time $t$, $t\geq 0$, the graph affinity between two vertices $i$ and $j$ is 
\begin{align*}
\graphAff(i,j)=A_t^2(i,j),
\end{align*}
where $A_t$ is the adjacency matrix at time $t$.  This graph affinity represents the number of common connections between $i$ and $j$. 

\paragraph{Graph Evolution:} The graph evolves in the following manner.  A vertex $i$ is picked at random. For each other vertex $j$, a ``model'' score $s(i,j)$ is computed as 
\begin{align*}
s(i,j)=p(G_i,G_j)\exp(\lambda \graphAff(i,j))\exp(\mu \aff_{\textrm{member}}(i,j)),
\end{align*}
where $\aff_{\textrm{member}}(i,j)=-\norm{X_i-X_j}_2$. The scores are then normalized by their sum. A subset of vertices of size $D$ with the highest scores are then considered eligible for showing. Among the eligible vertices, a smaller subset of size $m$ of the top $D$ vertices (possibly identified after re-ranking to incorporate fairness constraints) is then ``shown''.  The fairness constraints used for re-ranking are described in detail below.  For the subset of size $m$ of members that are shown, an edge (or connection) in the graph is created at random with probability $p_{\textrm{CTR}}$ at $k$, where $p_{\textrm{CTR}}$ is the position bias at $k$ multiplied by a base probability, $p_{\textrm{base}}$, with $p_{\textrm{base}}=.1$. This corresponds to an edge being created if and only if the member initiating the query requests it (which happens with probability $p_{\textrm{CTR}}$ at $k$).

%

\paragraph{Parameter Settings: } 
We used $p_0= 0.65$, as the fraction of group membership in group 0. We set $\sigma^2=.1$, and in our simulations, the covariates 
are such that $X_i\sim {\cal N}(\mu_{G_i},\sigma^2 I)$, the (two) $\mu_{G_i}$'s are picked at random, with $\mu_{G_i}\sim {\cal N}(0,I_{d_{cov}})$, $d_{cov}=30$. Block model parameters were $p_{00}=.05$,$p_{11}=.04$,$p_{01}=p_{01}=.01$. By default, $p(G_i,G_j)=1$. Position bias at $k$ is $\posBias(k)=1/(1+\log k)$. We used a graph with $10^3$ vertices and ran the simulation for $1,000$ iterations/sessions. The primal reranker is considerably slower than doing either no reranking or using the dual trick. So we ran this simulation for only $100$ iterations. To understand the impact of refreshing the dual more often, we ran simulations with the dual being refreshed every 50 and then 20 iterations. 

In the simulations, we picked $\lambda$ and $\mu$ adaptively so the ratio of the two highest scores is equal to .9 (though this is a tuning parameter that could be changed). This was to avoid a potential problem with the increase in the number of connections, which could end up disproportionately weighing the graph affinity and resulting in e.g. the largest score being overwhelming large compared to all other scores. By default, $p(G_i,G_j)=1$, i.e. there is no systematic bias per group membership, besides the one induced by the structure of the graph and the difference in covariates. $D=250$ though our simulations are flexible and can of course incorporate other parameters. The number $m$ of members ``shown'' is $m=10$, unless otherwise noted.

In the simulation setup above, Group 1 is disadvantaged in terms of exposure because of its smaller size (and weaker connectivity) and the way the scores are computed, the difference in means between the two groups rendering each group less visible to the other. Within groups the simulation setup pretty much guarantees homogeneity. This is partly why we set $p(G_i,G_j)=1$ and not a different function of $G_i$ and $G_j$ to illustrate the fact that even without systematic group bias some of the issues we had raised were already observable, owing just to initial network structure. 
  

\paragraph{Utlity Details: }
In the simulations reported below, the dest utility is the exposure utility, i.e. destination $j$ shown at position $k\leq m$ has utility $\destUtility(j)=\posBias(k)$ and 0 if not exposed. The source side utility for source $i$ is 
\begin{align*}
\sourceUtility(i)=\sum_{1\leq k \leq m} s(i,j_{(k)})\posBias(k)
\end{align*}
where $s(i,j_{(k)})$ is the $k$-th largest score $s(i,j)$. 

All utilities are aggregated by summation across sessions.  To make comparisons across simulations below more meaningful, we set the seed before the network generation phase and then set the seed before each of the run of the different methods. This ensures that the networks on which the simulations were run were exactly the same, that the queries were initiated in the same order etc. This is also why when displaying the same number of slots the simulation results are \textbf{exactly} the same for the methods not using a re-reranker. 

\paragraph{Optimization Details: } 
The primal simulations were comparatively very slow - in line with our expectations. 
Finally, the settings used for the LP solver can (not so surprisingly) have some impact of the performance of the rerankers, so we used quite stringent tolerance results in the results presented below. For the equality of opportunity constraint we consider $|f_{k,k'}^\top P_s w| < \epsilon$, where $\epsilon$ is chosen as:
\begin{align*}
\epsilon = \frac{1}{|[m]_{odd}|} \sum_{k \in [m]_{odd}}\posBias(k) -  \frac{1}{|[m]_{even}|} \sum_{k \in [m]_{even}}\posBias(k)
\end{align*}
where $[m]_{odd}$ and $[m]_{even}$ are the odd and even index of $[m] = 1, \ldots, m$. For the dynamic constraint we choose $ |\tilde u_s P_s w| \leq 0.1$.

\subsection{Metric Details}
We give more context on the metrics we track in the simulation:
\begin{enumerate}
\item \textbf{Demographic Parity in Exposure}: ($\Delta$DP) We consider the $m$ members ``shown''. For each query we compute the average exposure of Group 0 and Group 1, accounting for position bias. And we compute the difference between the average exposure of Group 0 and that of Group 1. We then report the mean across all sessions. 
\item \textbf{Absolute Demographic Parity in Exposure:} $\Delta$|DP|: this is the same as above, except at the next-to-last step we take the absolute value of the average exposure between groups for each session. Then we report the mean across sessions. 
\item \textbf{Average Source Utility $G_0$}: $\bar{U}_{G_0}^{\source}$: compute the source utility for members of group 0 who initiated at least 1 query. Source utility is the sum of scores of members recommended to the source weighted by position bias. 
\item \textbf{Average Source Utility $G_1$}: $\bar{U}_{G_1}^{\source}$: compute the source utility for members of group 1 who initiated at least 1 query. 
\item \textbf{Ratio of Average Source Utilities $G_0G_1$}: 
\begin{align*}
r_{G_0G_1}^{\source}: = \frac{\bar{U}_{G_0}^{\source}}{\bar{U}_{G_0}^{\source} + \bar{U}_{G_1}^{\source}}
\end{align*}
\item \textbf{Percentage Total Source Utility In $G_0$}: $PT_{G_0}^{\source}$:  computes the sum of source utilities (with at least 1 query) in $G_0$ and divide this by total source Utility (for members having initiated at least 1 query)
\item \textbf{Average Destination Utility $G_0$}: $\bar{U}_{G_0}^{dest}$:  Average dest Utility for members of group 0 having been recommended in at least 1 query. Dest utility for an individual member is measured by the position bias at the position they are being recommended at. 
\item \textbf{Average Destination Utility $G_1$}: $\bar{U}_{G_1}^{dest}$: average dest Utility for members of group 1 having been recommended in at least 1 query. 
\item \textbf{Ratio of Average Destination Utilities $G_0G_1$}: 
\begin{align*}
r_{G_0G_1}^{dest}: = \frac{\bar{U}_{G_0}^{dest}}{\bar{U}_{G_0}^{dest} + \bar{U}_{G_1}^{dest}}
\end{align*}
\item \textbf{Percentage Total Destination Utility In $G_0$}: $PT_{G_0}^{dest}$: this is the sum of dest Utility for members in Group 0 having been recommended at least once divided the total dest Utility for members having been recommended at least once. 
\end{enumerate}

\subsection{Full Results}
We now state the full results of the simulation results with respect to the different metrics as stated above.

\begin{table}[!h]
\centering
\begin{tiny}
\begin{tabular}{|c|c|c|c|c|c|c|c|c|c|c|}\hline
Method & $\Delta$DP & $\Delta$|DP| & $\bar{U}_{G_0}^{\source}$ & $\bar{U}_{G_1}^{\source}$& $r_{G_0G_1}^{\source}$&$PT_{G_0}^{\source}$&$\bar{U}_{G_0}^{dest}$&$\bar{U}_{G_1}^{dest}$&$r_{G_0G_1}^{dest}$&$PT_{G_0}^{dest}$\\ [1ex]
\hline\hline
No Reranker &-0.0216&0.1964&0.0114&0.0110&0.5101&0.6574&7.5894&5.8574&0.5644&0.7051 \\ \hline
Primal&0.0179&0.1671&0.0073&0.0073&0.5024&0.6349&1.2016&1.5402&0.4382&0.5371\\ \hline
Dual No Dynamic &0.0055&0.2126&0.0115&0.0110&0.5101&0.6574&6.3920&7.9805&0.4447&0.5861\\ \hline
Dual Dynamic&0.0249&0.2009&0.01147&0.0110&0.5100&0.6574&6.2242&8.2771&0.4292&0.5707\\ \hline
\end{tabular}
\end{tiny}
\vspace{0.1cm}
\caption{Here we have $1000$ members, $1000$ iterations, $m=10$ slots, and $50$ epochs in dual refresh.}
\label{tab:simulation1}
\end{table}

\begin{table}[!h]
\centering
\begin{tiny}
\begin{tabular}{|c|c|c|c|c|c|c|c|c|c|c|}\hline
Method & $\Delta$DP & $\Delta$|DP| & $\bar{U}_{G_0}^{\source}$ & $\bar{U}_{G_1}^{\source}$& $r_{G_0G_1}^{\source}$&$PT_{G_0}^{\source}$&$\bar{U}_{G_0}^{dest}$&$\bar{U}_{G_1}^{dest}$&$r_{G_0G_1}^{dest}$&$PT_{G_0}^{dest}$\\ [1ex]
\hline\hline
No Reranker&-0.0073&0.1182&0.0176&0.01689&0.5101&0.6574&11.2915&8.7335&0.5639&0.7059\\ \hline
Primal&0.0110&0.1002&0.0112&0.0111&0.5024&0.6348&1.3247&1.8482&0.4175&0.5269\\ \hline
Dual No Dynamic&0.0122&0.1090&0.0176&0.0169&0.5100&0.6573&8.9838&13.0638&0.4075&0.5537\\ \hline
Dual Dynamic&0.0156&0.1010&0.0176&0.0169&0.5100&0.6573&8.7293&13.5090&0.3925&0.5372\\ \hline
\end{tabular}
\end{tiny}
\vspace{0.1cm}
\caption{Here we have $1000$ members, $1000$ iterations, $m=20$ slots, and $50$ epochs in dual refresh.}
\label{tab:simulation2}
\end{table}

\begin{table}[!h]
\centering
\begin{tiny}
\begin{tabular}{|c|c|c|c|c|c|c|c|c|c|c|}\hline
Method & $\Delta$DP & $\Delta$|DP| & $\bar{U}_{G_0}^{\source}$ & $\bar{U}_{G_1}^{\source}$& $r_{G_0G_1}^{\source}$&$PT_{G_0}^{\source}$&$\bar{U}_{G_0}^{dest}$&$\bar{U}_{G_1}^{dest}$&$r_{G_0G_1}^{dest}$&$PT_{G_0}^{dest}$\\ [1ex]
\hline\hline
No Reranker &-0.0073&0.1182&0.0176&0.0169&0.5101&0.6574&11.2915&8.7335&0.5639&0.7059\\ \hline
Primal&0.0110&0.1002&0.0112&0.0111&0.5024&0.6348&1.3246&1.8482&0.4175&0.5269\\ \hline
Dual no Dynamic&0.0177&0.1106&0.0176&0.0169&0.5100&0.6573&8.4811&13.9792&0.3776&0.5210\\ \hline
Dual with Dynamic&0.0186&0.1090&0.0176&0.0169&0.5099&0.6573&8.3701&14.0656&0.3731&0.5167\\ \hline
\end{tabular}
\end{tiny}
\vspace{0.1cm}
\caption{Here we have $1000$ members, $1000$ iterations, $m=20$ slots, and $20$ epochs in dual refresh.}
\label{tab:simulation3}
\end{table}

\subsection{Bootstrap Results}
We also repeat each of the following above results $100$ times with varying seed to estimate the variance of each of metrics. The detailed table result is given below with the mean and 95\% error estimates.

\begin{table}[!h]
\centering
\begin{tiny}
\begin{tabular}{|c|c|c|c|c|c|c|c|c|c|c|}\hline
Method & $\Delta$DP & $\Delta$|DP| & $\bar{U}_{G_0}^{\source}$ & $\bar{U}_{G_1}^{\source}$& $r_{G_0G_1}^{\source}$&$PT_{G_0}^{\source}$&$\bar{U}_{G_0}^{dest}$&$\bar{U}_{G_1}^{dest}$&$r_{G_0G_1}^{dest}$&$PT_{G_0}^{dest}$\\ [1ex]
\hline\hline
\multirow{2}{*}{No Reranker}     & 0.0000196 & 0.1876794 & 0.0114761 & 0.0110283 & 0.5099510 & 0.6572877 & 7.1818291 & 6.5941001 & 0.5214912 & 0.6709937\\
 & ($\pm$1.53e-3) & ($\pm$1.23e-3) & ($\pm$1.43e-6) & ($\pm$1.25e-6) & ($\pm$3.85e-5) & ($\pm$3.47e-5) & ($\pm$3.41e-2) & ($\pm$5.89e-2) & ($\pm$3.33e-3) & ($\pm$2.93e-3)\\ \hline
 \multirow{2}{*}{Primal} & 0.0286859 & 0.1837153 & 0.0073293 & 0.0072788 & 0.5017288 & 0.6342587 & 1.2185111 & 1.5345218 & 0.4426842 & 0.5331251\\
 & ($\pm$2.15e-3) & ($\pm$2.10e-3) & ($\pm$1.58e-6) & ($\pm$1.43e-6) & ($\pm$6.90e-5) & ($\pm$6.40e-5) & ($\pm$7.97e-3) & ($\pm$1.16e-2) & ($\pm$2.26e-3) & ($\pm$1.06e-3)\\ \hline
Dual & 0.0450135 & 0.2051839 & 0.0114708 & 0.0110255 & 0.5098977 & 0.6572397 & 6.0227132 & 8.6437419 & 0.4109908 & 0.5569184\\
No Dynamic & ($\pm$2.92e-3) & ($\pm$2.05e-3) & ($\pm$1.46e-6) & ($\pm$1.27e-6) & ($\pm$3.92e-5) & ($\pm$3.53e-5) & ($\pm$5.08e-2) & ($\pm$9.33e-2) & ($\pm$4.62e-3) & ($\pm$4.91e-3)\\ \hline
Dual & 0.0461693 & 0.2047534 & 0.0114707 & 0.0110256 & 0.5098946 & 0.6572369 & 6.0011009 & 8.6903315 & 0.4088543 & 0.5547041\\
With Dynamic & ($\pm$2.92e-3) & ($\pm$2.08e-3) & ($\pm$1.48e-6) & ($\pm$1.28e-6) & ($\pm$3.95e-5) & ($\pm$3.56e-5) & ($\pm$5.17e-2) & ($\pm$9.71e-2) & ($\pm$4.75e-3) & ($\pm$5.00e-3)\\ \hline
 \end{tabular}
\end{tiny}
\vspace{0.1cm}
\caption{$m=10$ slots and $50$ epochs in dual refresh.}
\label{tab:simulationBoot1}
\end{table}

\begin{table}[!h]
\centering
\begin{tiny}
\begin{tabular}{|c|c|c|c|c|c|c|c|c|c|c|}\hline
Method & $\Delta$DP & $\Delta$|DP| & $\bar{U}_{G_0}^{\source}$ & $\bar{U}_{G_1}^{\source}$& $r_{G_0G_1}^{\source}$&$PT_{G_0}^{\source}$&$\bar{U}_{G_0}^{dest}$&$\bar{U}_{G_1}^{dest}$&$r_{G_0G_1}^{dest}$&$PT_{G_0}^{dest}$\\ [1ex]
\hline\hline
\multirow{2}{*}{No Reranker}     & 0.0020038 & 0.1097322 & 0.0175915 & 0.0169085 & 0.5098985 & 0.6572404 & 10.7350035 & 9.8549059 & 0.5215575 & 0.6700181\\
 & ($\pm$9.01e-4) & ($\pm$6.86e-4) & ($\pm$2.11e-6) & ($\pm$1.81e-6) & ($\pm$3.65e-5) & ($\pm$3.29e-5) & ($\pm$5.11e-2) & ($\pm$9.29e-2) & ($\pm$3.49e-3) & ($\pm$3.02e-3)\\ \hline
 \multirow{2}{*}{Primal} & 0.0107691 & 0.1045706 & 0.0112299 & 0.0111545 & 0.5016847 & 0.6342178 & 1.3473105 & 1.8612044 & 0.4199544 & 0.5273710\\
 & ($\pm$1.47e-3) & ($\pm$1.28e-3) & ($\pm$2.32e-6) & ($\pm$2.06e-6) & ($\pm$6.36e-5) & ($\pm$5.91e-5) & ($\pm$7.65e-3) & ($\pm$1.10e-2) & ($\pm$1.86e-3) & ($\pm$7.04e-4)\\ \hline
Dual & 0.0182423 & 0.1104472 & 0.0175825 & 0.0169060 & 0.5098076 & 0.6571585 & 8.7662834 & 13.4431304 & 0.3950420 & 0.5437587\\
No Dynamic & ($\pm$1.17e-3) & ($\pm$9.75e-4) & ($\pm$2.19e-6) & ($\pm$1.90e-6) & ($\pm$3.73e-5) & ($\pm$3.36e-5) & ($\pm$7.50e-2) & ($\pm$1.39e-1) & ($\pm$4.51e-3) & ($\pm$4.71e-3)\\ \hline
Dual & 0.0172247 & 0.1087706 & 0.0175828 & 0.0169063 & 0.5098077 & 0.6571586 & 8.7509092 & 13.4787915 & 0.3939329 & 0.5426752\\
With Dynamic & ($\pm$1.09e-3) & ($\pm$8.26e-4) & ($\pm$2.18e-6) & ($\pm$1.90e-6) & ($\pm$3.78e-5) & ($\pm$3.41e-5) & ($\pm$7.17e-2) & ($\pm$1.28e-1) & ($\pm$4.21e-3) & ($\pm$4.37e-3)\\ \hline \end{tabular}
\end{tiny}
\vspace{0.1cm}
\caption{$m=20$ slots and $50$ epochs in dual refresh.}
\label{tab:simulationBoot2}
\end{table}

\begin{table}[!h]
\centering
\begin{tiny}
\begin{tabular}{|c|c|c|c|c|c|c|c|c|c|c|}\hline
Method & $\Delta$DP & $\Delta$|DP| & $\bar{U}_{G_0}^{\source}$ & $\bar{U}_{G_1}^{\source}$& $r_{G_0G_1}^{\source}$&$PT_{G_0}^{\source}$&$\bar{U}_{G_0}^{dest}$&$\bar{U}_{G_1}^{dest}$&$r_{G_0G_1}^{dest}$&$PT_{G_0}^{dest}$\\ [1ex]
\hline\hline
\multirow{2}{*}{No Reranker}  & 0.0020038 & 0.1097322 & 0.0175915 & 0.0169085 & 0.5098985 & 0.6572404 & 10.7350035 & 9.8549059 & 0.5215575 & 0.6700181\\
 & ($\pm$9.01e-4) & ($\pm$6.86e-4) & ($\pm$2.11e-6) & ($\pm$1.81e-6) & ($\pm$3.65e-5) & ($\pm$3.29e-5) & ($\pm$5.11e-2) & ($\pm$9.29e-2) & ($\pm$3.49e-3) & ($\pm$3.02e-3)\\ \hline
\multirow{2}{*}{Primal} & 0.0107691 & 0.1045706 & 0.0112299 & 0.0111545 & 0.5016847 & 0.6342178 & 1.3473105 & 1.8612044 & 0.4199544 & 0.5273710\\
 & ($\pm$1.47e-3) & ($\pm$1.28e-3) & ($\pm$2.32e-6) & ($\pm$2.06e-6) & ($\pm$6.36e-5) & ($\pm$5.91e-5) & ($\pm$7.65e-3) & ($\pm$1.10e-2) & ($\pm$1.86e-3) & ($\pm$7.04e-4)\\ \hline
Dual & 0.0224202 & 0.1102967 & 0.0175802 & 0.0169053 & 0.5097851 & 0.6571382 & 8.2356443 & 14.4325451 & 0.3634357 & 0.5092317\\
No Dynamic & ($\pm$1.02e-3) & ($\pm$7.34e-4) & ($\pm$2.15e-6) & ($\pm$1.85e-6) & ($\pm$3.74e-5) & ($\pm$3.37e-5) & ($\pm$4.94e-2) & ($\pm$8.90e-2) & ($\pm$2.76e-3) & ($\pm$3.00e-3)\\ \hline
Dual & 0.0211929 & 0.1086926 & 0.0175805 & 0.0169055 & 0.5097863 & 0.6571393 & 8.2246019 & 14.4590305 & 0.3626681 & 0.5085179\\
With Dynamic & ($\pm$9.40e-4) & ($\pm$7.08e-4) & ($\pm$2.14e-6) & ($\pm$1.86e-6) & ($\pm$3.72e-5) & ($\pm$3.35e-5) & ($\pm$4.20e-2) & ($\pm$7.59e-2) & ($\pm$2.35e-3) & ($\pm$2.53e-3)\\ \hline

  \end{tabular}
\end{tiny}
\vspace{0.1cm}
\caption{$m=20$ slots and $20$ epochs in dual refresh.}
\label{tab:simulationBoot2}
\end{table}

Note that over 100 iterations we get consistent results. That is source side utilities are usually automatically balanced. The dest side utilities improve in the case when $m$ increases from 10 to 20. Also, we fast dual refresh, we are getting better results, by sacrificing computation time.

\end{document}